\documentclass[letterpaper, 10 pt, conference]{ieeeconf}  
\IEEEoverridecommandlockouts 	
\usepackage{graphicx}		
\usepackage{algorithm}		
\usepackage{algpseudocode}	
\usepackage{amsmath} 		
\usepackage{amssymb}		
\usepackage{array}			
\usepackage{bm}			
\usepackage{comment}		
\usepackage{multirow}		
\usepackage{multicol}		
\usepackage{subfigure}		
\usepackage{cancel}			
\usepackage{hyperref} 		
\usepackage{fancyhdr} 		
\usepackage{lastpage} 		
\usepackage{xcolor}			
\usepackage[framed,numbered,autolinebreaks,useliterate]{mcode}	
\usepackage{dsfont}			
\usepackage[font={small}]{caption}

\definecolor{b}{rgb}{0.0, 0.0, 0.6}
\definecolor{g}{rgb}{0.0, 0.6, 0.0}
\definecolor{r}{rgb}{0.9, 0.5, 0.2}

\graphicspath{{./Figures/}}

\title{\LARGE \bf \textit{Out-of-focus}: Learning Depth from Image Bokeh for Robotic Perception}
\author{Eric Cristofalo and Zijian Wang\\
	CS 229 Project Report \\
	December 16, 2016
\thanks{The authors are affiliated with the Multi-robot Systems Lab (MSL) in the Department of Aeronautics and Astronautics, Stanford University, United States.
    \texttt{\small \{ecristof, zjwang\}@stanford.edu}.}
\thanks{}}

\begin{document}
\maketitle

\thispagestyle{plain} 
\pagestyle{plain}

\begin{abstract}
In this project, we propose a novel approach for estimating depth from RGB images. Traditionally, most work uses a single RGB image to estimate depth, which is inherently difficult and generally results in poor performance -- even with thousands of data examples. In this work, we alternatively use multiple RGB images that were captured while changing the focus of the camera's lens. This method leverages the natural depth information correlated to the different patterns of clarity/blur in the sequence of focal images, which helps distinguish objects at different depths. Since no such data set exists for learning this mapping, we collect our own data set using customized hardware. We then use a convolutional neural network for learning the depth from the stacked focal images. Comparative studies were conducted on both a standard RGB-D data set and our own data set (learning from both single and multiple images), and results verified that stacked focal images yield better depth estimation than using just single RGB image.
\end{abstract}

\section{Introduction}
\label{sec:introduction}

Estimating pixel depth from raw images is a difficult task that is often solved using multiple salient images or incorporating active sensors, e.g., laser rangefinders. Unfortunately, vision-based methods are computationally expensive due to the significant processing required to match features and build detailed maps. Furthermore, they rarely take advantage of another useful feature of the optical lens-sensor construction -- the out-of-focus aesthetic, or \textit{bokeh}, of an image. Humans and animals are especially good at using out-of-focus image information to perceive three-dimensional aspects of the world, e.g. peripheral vision for obstacle avoidance or depth-of-field in photography. In fact, a significant portion of our view is often out-of-focus, allowing us to concentrate on what we deem important at the moment. 

There have been extensive studies on the quantification of focus (or out-of-focus) in a image to determine the corresponding depth. Most notably, the authors of~\cite{pertuz2013analysis} compared 36 measures of focus including Laplacian-based, wavelet-based, fourier-based, and various other methods. The field of confocal microscopy also utilizes these focus measures to create high-fidelity maps of microscopic terrain~\cite{nayar1994shape}. However, all measures suffer from lack of texture or contrast and ultimately limit the usefulness of out-of-focus images for autonomous robotics applications. We alternatively learn the mapping from out-of-focus images to dense pixel depth maps in this work. 

We use supervised learning for this problem. More specifically, we take as an input a \textit{focal stack} of varied focus images and predict a per-pixel depth map correspondence of the scene. To to this, we use labeled data from an RGB-D camera along with the corresponding focal stack of images. The output from the infrared camera can be treated as the ground truth depth information, or data labels, while out-of-focus images from the RGB camera are the features. We utilize neural networks for this supervised learning task. Neural networks, especially deep/convolutional networks, have gained great success recently in image and speech recognition. The advantage of neural network is that we do not need to hand code the feature/kernel, which allows for more possibility and flexibility for learning. 

Although there exists many RGB-D data sets~\cite{lai2011large, sturm2012benchmark, shotton2013scene}, this task is challenging because we require out-of-focus image stacks and their corresponding ground truth depth maps all captured from a static camera assembly. We therefore utilize tools from the Multi-robot Systems Lab to obtain this data set to test our method. In addition, since our end application is robotics, we also utilize a camera with electronically controllable focus. The data collection, image processing, and prediction can be run in real time, allowing the future use of a mobile robot platform to navigate autonomously while avoiding obstacles. 

\subsection{Related Work}
Depth maps are of great important in computer vision. Traditionally, there is a lot of research studying how to estimate depth using just a single RGB image. In \cite{eigen2014depth}, the depth was estimated by a multi-scale perspective, namely, by combining coarse global estimation and local fine estimations. Convolutional Neural Fields (CNF) were proposed in \cite{liu2015learning} by exploiting the fact that depth is continuous and therefore can be described as fields. In \cite{choi2015depth}, the authors argued that estimating depth from a single monocular image is an ill-posed problem. They tackled the problem by accounting for a collection of consecutively taken RGB-D images and incorporated the spatial correlation into the depth estimation. Further, the literature on semantic segmentation \cite{long2015fully} is also relevant to our work because a continuous piece of depth information usually indicates a single object. 

On the other hand, very few researchers have leveraged a camera's variable focal length in learning depth, which is used in our approach. In terms of estimating depth from out-of-focus images, Petland produced the first notable example in 1987~\cite{pentland1987new}. The authors of~\cite{lin2013absolute} used single de-focused image to estimate depth. One of the more impressive examples~\cite{suwajanakorn2015depth} attempted to optimize a cost function for pixel depth by accounting for image alignment and over all smoothness. All of these works still rely on a robust method of quantifying focus, which is difficult to generalize for any scene or lighting condition. Our method, on the other hand, does not require the definition of any such measure and can be learned from data. 

\section{Problem Statement}
\label{sec:problemStatement}

The goal of this project is to learn depth from a sequence of out-of-focus image using convolutional neural networks due to their popularity and success within computer vision problems. The out-of-focus images represent a camera capturing the same, static scene while varying its focal length. This is an interesting problem because, unlike learning from single RGB images, our out-of-focus images inherently contain some notion of individual pixel-wise depth. Therefore, a secondary goal of this project is to explicitly show the benefit of learning from focus compared to simply RGB images. 

We have created a data set for learning this information that includes 3674 data examples in total. The data was collected using a mobile, multiple camera experimental setup including an RGB-D sensor for labeled data and a standard webcam with variable focus for the out-of-focus images (Section~\ref{sec:dataSet}).
We have implemented a convolutional neural network in Keras using TensorFlow for learning depth from images (Section~\ref{sec:neuralNetwork}). This flexible network architecture allows us to fairly compare three main experiments:
\begin{itemize}
    \item[\textbf{1)}] learning depth from a single RGB image from the NYU data set, 
    \item[\textbf{2)}] learning depth from a single RGB image from our collected data set, and
    \item[\textbf{3)}] learning depth from a focal stack of RGB images from our collected data set. 
\end{itemize}
We compare our findings and results in Section~\ref{sec:results}.

\section{Depth-from-Focus Data Set}
\label{sec:dataSet}

\begin{figure}
\centering
        \subfigure[Laptop running ROS\label{fig:ex_setup1}]{\includegraphics[width = 0.49\linewidth]{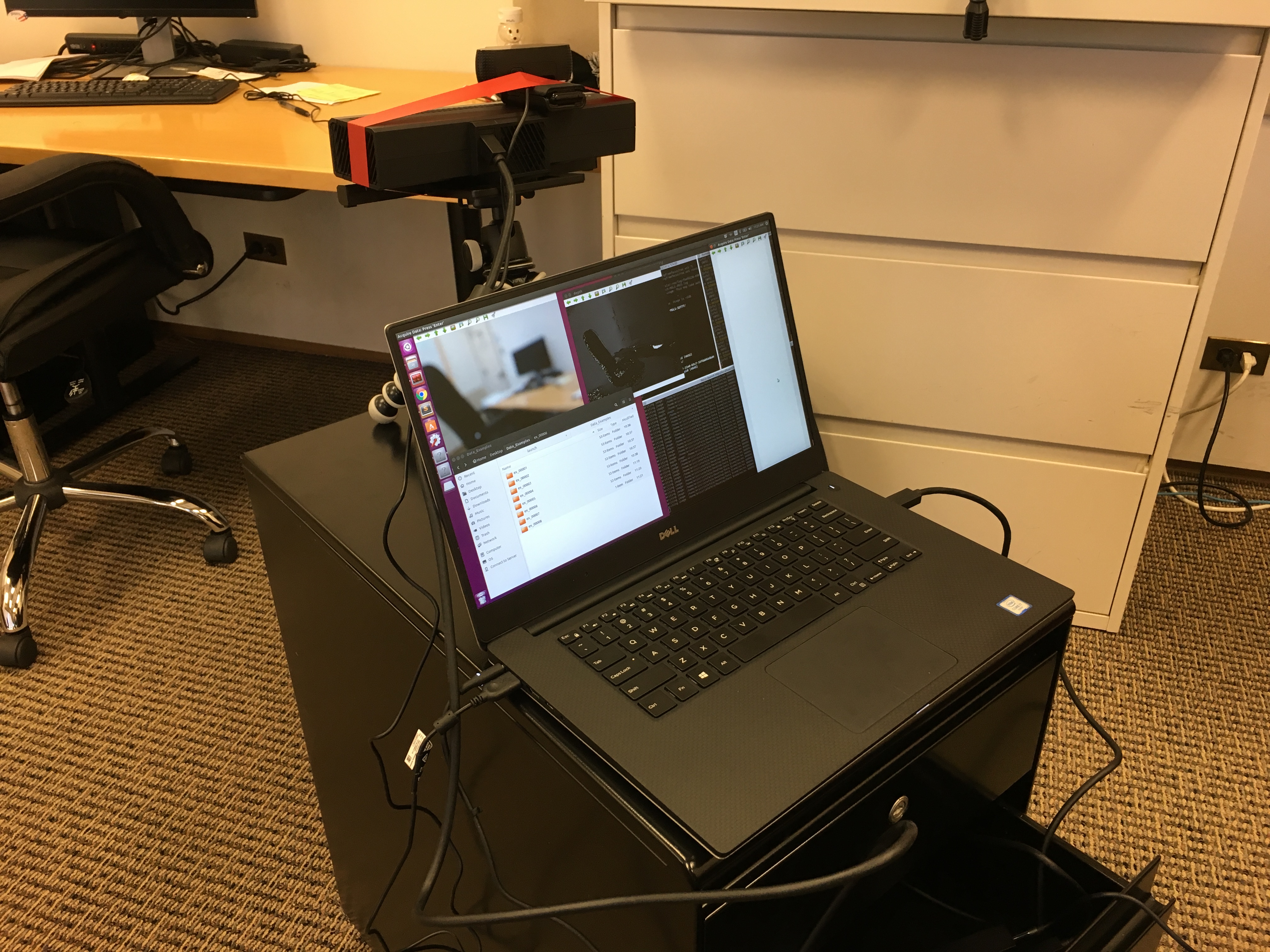}}
        \subfigure[Two camera setup\label{fig:ex_setup2}]{\includegraphics[width = 0.49\linewidth]{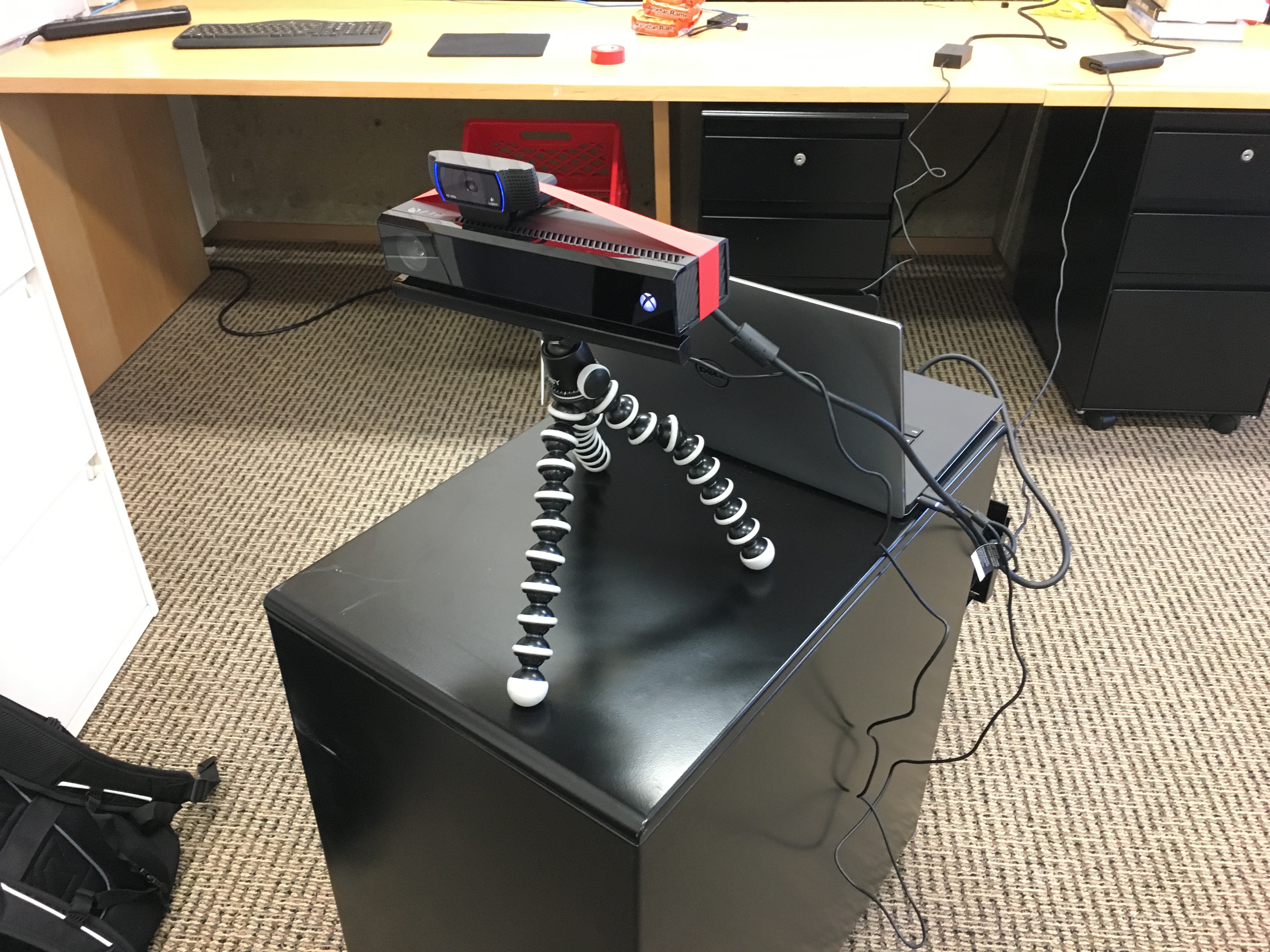}}
        \caption{Mobile experimental setup for data set collection includes a laptop running Ubuntu and ROS (left), and two cameras (right): Logitech C920 usb webcam and Microsoft Kinect}
\label{fig:03_exp_setup}
\end{figure}

The labeled training data was collected using two different sensors: a Microsoft Kinect V2~\cite{Kinect} and a Logitec C920 webcam~\cite{Logitech} (See Fig.~\ref{fig:03_exp_setup}) and represents a collection of scenes expected in a typical office environment. 
The Kinect includes a 1080p camera and a infrared depth sensor with a volume covering approximately 0.4 to 4 meters of depth range. For visualization, in the remainder of this paper the depth images from the Kinect are scaled to RGB images by assigning the minimum distance (0.4m) to the color blue and the maximum distance (4m) to the color red. The Logitech webcam also captures 1080p images and includes an autofocusing mechanism capable of up to 256 autofocusing stops. 
These sensors were chosen because of their reasonable price and sufficient performance for collected the focal stack and labeled images. 

\begin{figure}
\centering
    \includegraphics[width=1\linewidth]{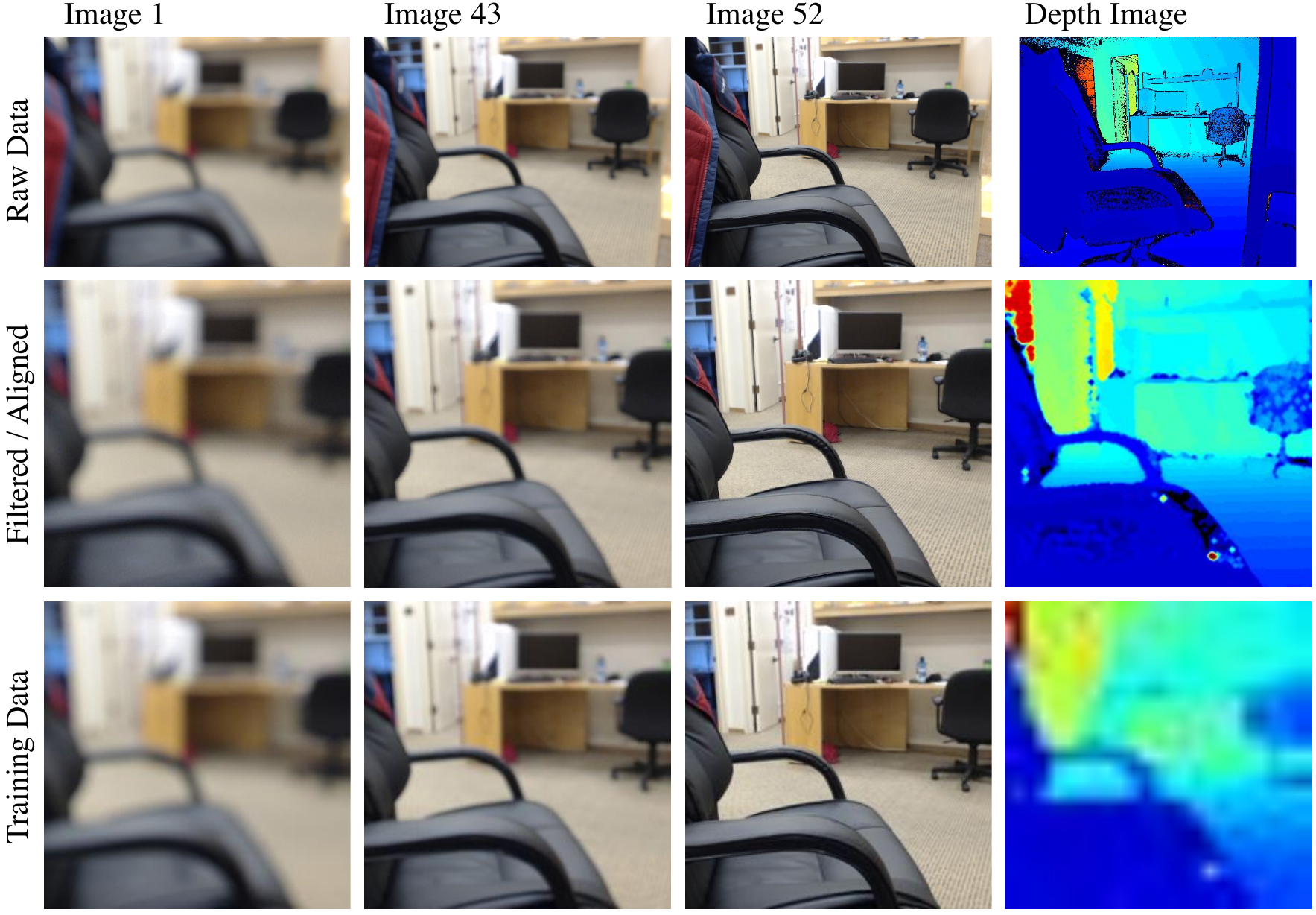}
    \caption{Example data from training set. Sample of three raw focal stack images (52 total images with varying focal depth) and corresponding raw depth map (top row). Filtered and aligned image data (middle row). Final training images and labeled data (bottom row). We collected 344 unique training examples in total. }
\label{fig:03_data_example}
\end{figure}

An example of the raw data captured with these sensors is shown in the first row of Fig.~\ref{fig:03_data_example}. The data consists of $N=52$ images with varying \textit{focal depth}, which is defined as the distance from the center of the lens to the image sensor (measure in mm). These images represent the same scene acquired with a different focal depth that was varied linearly through the allowable range of the camera (0-255 in this case). Associated to this stack of images -- denoted as the \textit{focal stack} -- is the labeled \textit{depth image} acquired by the Kinect. The focal stack is comprised of 52 $640 \times 480 \times 3$ 8-bit images and the depth image is a $512 \times 424 \times 1$ 32-bit image. 

The collection of data was facilitated with the Robot Operating System (ROS)~\cite{Quiqley2009ROS} which synchronized the data collection from the two sensors. We utilized the OpenCV libraries~\cite{bradski2008learning} to acquire the images and save them with their corresponding labeled depth image. The two sensors were carefully attached to ensure their relative position remained constant for accurate and consistent data alignment. 

The raw data was pre-processed before training in order to filter and register the images from the two cameras (see results of filtering and registration in the second row of Fig.~\ref{fig:03_data_example}). The filtering on the depth map is required because the raw depth data is corrupted by noise from either reflective surfaces or external infrared light interference. We use a conventional erosion-dilation and Gaussian bluring pipeline to clean the depth map. We then align the images from the two cameras to account for their 3D translational offset -- defined by the webcam position approximately 0.05 meters above the Kinect's depth sensor. Using webcam's camera calibration matrix (which we found using OpenCV), we warp the 3D point cloud from the Kinect into the camera's coordinate frame using a perspective transformation. We finally crop and resize the images yielding the final training data (third row of Fig.~\ref{fig:03_data_example}). The final focal images are of size $224 \times 224 \times 3$ and the depth map is of size $24 \times 24 \times 1$. 

\begin{figure}
\centering
    \subfigure{\includegraphics[width = 0.24\linewidth]{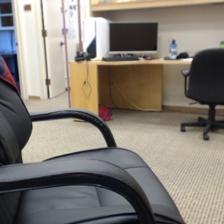}}
    \subfigure{\includegraphics[width = 0.24\linewidth]{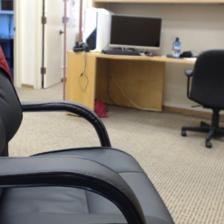}}
    \subfigure{\includegraphics[width = 0.24\linewidth]{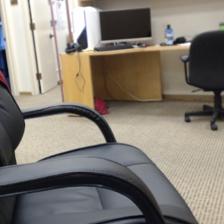}}
    \subfigure{\includegraphics[width = 0.24\linewidth]{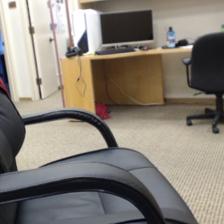}}
    \caption{Example data from training set that has been perturbed using a random 3D perspective transformation. We obtained 3674 total training examples after applying these perturbations. }
\label{fig:03_rotation_example}
\end{figure}

In total, we captured 344 unique examples of the office scene using this method. We increased the number of training data examples to 3674 by randomly perturbing each original example ten times. This was achieved by transforming each example with a finite random 3D perspective transformation, resulting in the example images shown in Fig~\ref{fig:03_rotation_example}. 

\section{Methods}
\label{sec:neuralNetwork}
We use a convolutional neural network (CNN) for learning the depth from the RGB images due to its recent success in image recognition. A CNN performs a convolution operation on the RGB image, which iterates through small patches of neighboring pixels and then performs a kernel mapping to extract features. This is suitable for our application since judging the clarity/blur of object will require considering the neighboring pixels. Another advantage of CNN is that it naturally handles the depth of the input, which is convenient and efficient for stacked images. For typical use of a CNN, the single input image has depth 3 corresponding to the R, G, and B channels. In the multi-image case, we can stack $N$ images together and use a depth of $3N$. 

\subsection{Network Architecture}
\begin{figure}[h]
\centering
        \includegraphics[width = 1\linewidth]{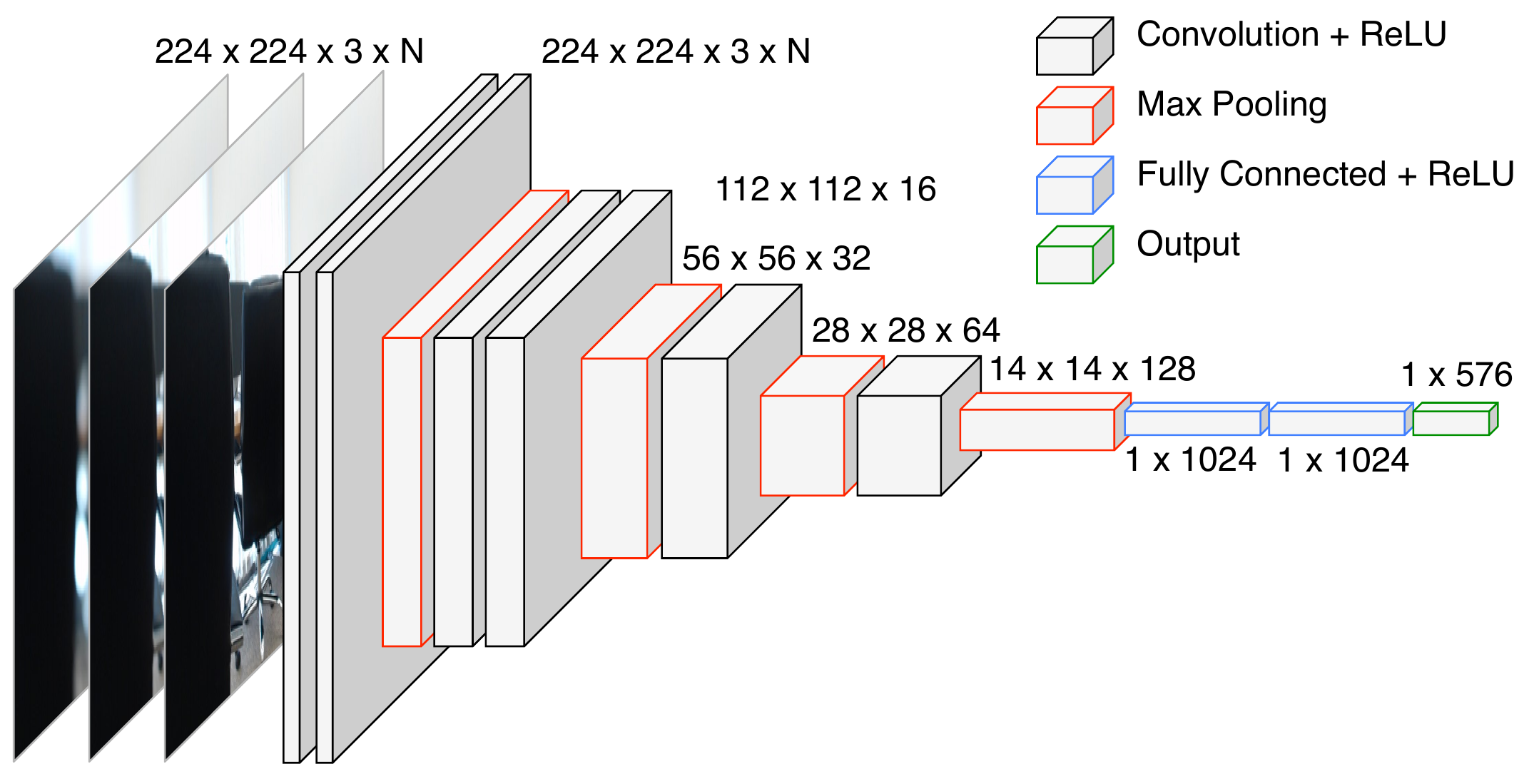}
        \caption{The architecture of the convolutional neural network used for training}
\label{fig:convArch}
\end{figure}

The structure of our CNN is visualized in Figure \ref{fig:convArch}. It contains 6 convolutional layers with 3x3 stride and ReLU activation, 4 max pooling layers, and 2 fully connected layers. Among these different components, convolutional layers take care of extracting spatial features through filtering on patches of pixels; max pooling layers effectively shrink the size of the image while retaining the critical information; the fully connected layers are responsible for synthesizing the final output. To ensure performance, our network structure inherits some design philosophy and suggestions from the well-known VGG network \cite{simonyan2014very}, including small stride size and the cascading of multiple convolution and max pooling layers. However, we have not gone as deep as the VGG network in order to avoid overfitting since we currently do not have access to a sufficiently large data set. 

\subsection{Loss Function and Training}
We use mean square error as the loss function. Concretely, the output of our CNN is a $576 \times 1$ vector, denoted as $x_p$. Note that $x_p$ corresponds to a flattened $24 \times 24$ estimated depth image. Also refer to the labeled ground truth depth image as a flattened vector $x$. Then the mean square error can be written as 
$$
e = \frac{1}{576} \sum_{i=1}^{576} (x(i) - x_p(i))^2,
$$
where the parenthesis of $x(i)$ denote the indexing of individual pixels. For training, we randomly split data into 85\% training examples and 15\% testing examples. SGD is used to carry out the backpropagation in order to optimize the weights in the network. Before training, the images are also shuffled in order to remove the potential undesired correlation between the training sequences.

\subsection{Comparative Study}

In order to verify our argument that using stacked focal images is advantageous, we compare the the corresponding performance using the same network structure, and only change the format of the input.

\textbf{Exp 1.} In the first experiment, we train our neural network on a well-known RGB-D data set called NYU Depth V2 \cite{eigen2014depth}. The input from this data set is single RGB image and the output is a depth image collected from a Microsoft Kinect sensor, which we use as the ground truth for supervised learning. The purpose of testing on the well-known data set is to expose our CNN implementation to a large data set (from which we extracted 11670 examples), and generally verify the effectiveness of our algorithm. 

\textbf{Exp 2.} For a fairer comparison, we further attempt to learn depth from single RGB image using examples in our own data set. As argued before, we expect this to perform poorly due to the lack of focal information. The single RGB image is taken as the $N^{\mathrm{th}}$ image of the focal stack, or image 52, because it includes the most in-focus information in the Kinect's usable volume

\textbf{Exp 3.} In this main experiment, we use stacked images with equally spaced focal lengths as the training input. This is in contrast to Exp 2, where only the last image with infinitely long focal length is used for depth estimation. Other than the difference of the input, other conditions of the training are exactly the same as Exp 2. We are therefore interested in observing how much improvement can it have compared to Exp 2. 

\section{Results and Discussion}
\label{sec:results}
We implemented the CNN in Keras \cite{chollet2015keras} with Tensorflow \cite{abadi2016tensorflow} background. All of our source code as well as some of the preprocessed training data are hosted online at \url{https://bitbucket.org/cs_229/learningfromfocus}. For the NYU Depth data set, we used 11670 data examples. For our own data set, we acquired 3674 samples (each sample contains 52 focal images and 1 Kinect depth output). All the data are split into 85\% for training, and 15\% testing examples for checking the performance. We do not utilize cross validation here due to time constraints, since the training usually takes more than five hours on a 12-core, 64-GB memory desktop workstation. Also, the training data is fed for training in a 32-example mini-batch in order to reduce the variance of the SGD update. 

\begin{figure}[h]
\centering
        \includegraphics[width = 0.9\linewidth]{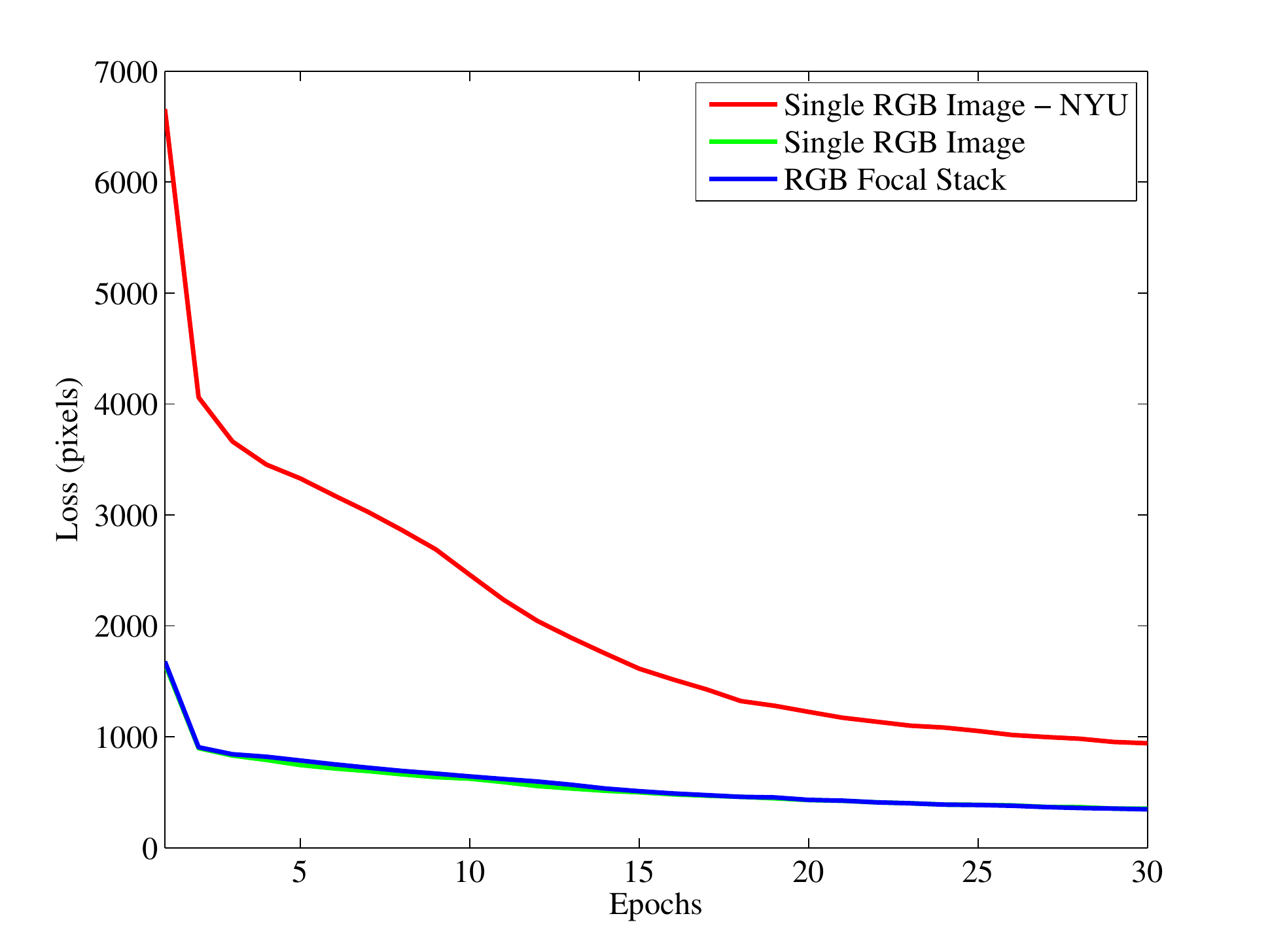}
        \caption{Training errors over time of the three experiments in Sec. \ref{sec:neuralNetwork}.}
\label{fig:LearningCurveLoss}
\end{figure}

The losses for the three experiments described in Sec. \ref{sec:neuralNetwork} during the training phase are shown in Figure \ref{fig:LearningCurveLoss}. It can be seen in all cases that the training errors converge after 30 epochs. The training error on the NYU data set is slightly higher, potentially because the NYU data set is much larger and has richer types of scenes. The two experiments on our own data set have much closer training errors. However, Exp 3 (the one using stacked focal images) has slightly better performance by the end of the training, in terms of both training error and testing error, as summarized in Table I. This result verifies our motivation that using stacked focal images is indeed better than just using one single RGB image. 

\begin{table}
\label{tab:trainTestError}
\begin{center}
\begin{tabular}{|c|c|c|c|}
\hline 
Exp \#  &  1   &  2  & 3\\ 
  &  NYU   &  Ours, Single  & Ours, Stack\\ 
\hline
Train MSE &  941.06  &  352.73  & 345.23   \\
\hline
Train MAE &  22.93  &  12.92  & 12.79   \\
\hline
Test MSE &  698.99  &  355.27  & 335.88   \\
\hline
Test MAE &  19.00  &  12.53  & 12.11   \\
\hline
\end{tabular}
\end{center}
\caption{Training and testing errors. Error is the average per pixel. MSE = mean squared error. MAE = mean absolute error.}
\end{table}

\begin{figure}
\centering
\renewcommand{\thesubfigure}{}
    \subfigure{\includegraphics[width = 0.26\linewidth]{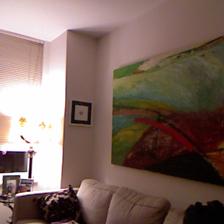}}
    \subfigure{\includegraphics[width = 0.26\linewidth]{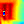}}
    \subfigure{\includegraphics[width = 0.26\linewidth]{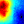}}
    \subfigure{\includegraphics[width = 0.26\linewidth]{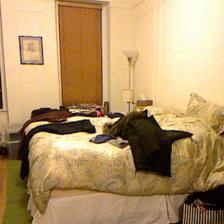}}
    \subfigure{\includegraphics[width = 0.26\linewidth]{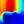}}
    \subfigure{\includegraphics[width = 0.26\linewidth]{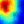}}
    \subfigure{\includegraphics[width = 0.26\linewidth]{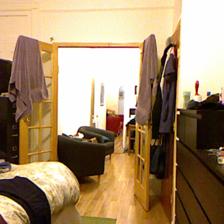}}
    \subfigure{\includegraphics[width = 0.26\linewidth]{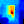}}
    \subfigure{\includegraphics[width = 0.26\linewidth]{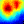}}
    \subfigure[Raw Image]{\includegraphics[width = 0.26\linewidth]{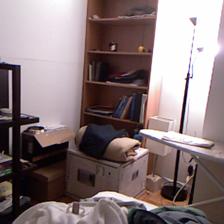}}
    \subfigure[Ground Truth]{\includegraphics[width = 0.26\linewidth]{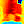}}
    \subfigure[Single Prediction]{\includegraphics[width = 0.26\linewidth]{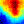}}
    \caption{Testing results on unseen data from NYU data set illustrating prediction from single images only. Each row represents an individual testing example. Each column represents the raw image (left), the ground truth depth map (middle), and the single image prediction (right), respectively}
\label{fig:05_nyu_results}
\end{figure}

\begin{figure}[h!]
\centering
\renewcommand{\thesubfigure}{}
    \subfigure{\includegraphics[width = 0.24\linewidth]{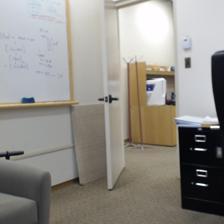}}
    \subfigure{\includegraphics[width = 0.24\linewidth]{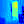}}
    \subfigure{\includegraphics[width = 0.24\linewidth]{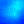}}
    \subfigure{\includegraphics[width = 0.24\linewidth]{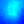}}
    \subfigure{\includegraphics[width = 0.24\linewidth]{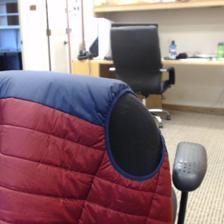}}
    \subfigure{\includegraphics[width = 0.24\linewidth]{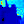}}
    \subfigure{\includegraphics[width = 0.24\linewidth]{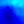}}
    \subfigure{\includegraphics[width = 0.24\linewidth]{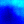}}
    \subfigure{\includegraphics[width = 0.24\linewidth]{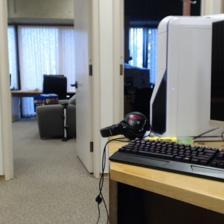}}
    \subfigure{\includegraphics[width = 0.24\linewidth]{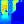}}
    \subfigure{\includegraphics[width = 0.24\linewidth]{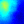}}
    \subfigure{\includegraphics[width = 0.24\linewidth]{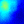}}
    \subfigure{\includegraphics[width = 0.24\linewidth]{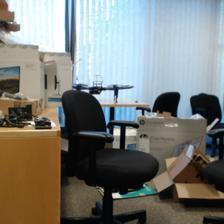}}
    \subfigure{\includegraphics[width = 0.24\linewidth]{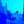}}
    \subfigure{\includegraphics[width = 0.24\linewidth]{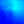}}
    \subfigure{\includegraphics[width = 0.24\linewidth]{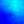}}
    \subfigure{\includegraphics[width = 0.24\linewidth]{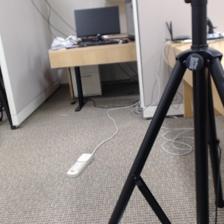}}
    \subfigure{\includegraphics[width = 0.24\linewidth]{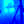}}
    \subfigure{\includegraphics[width = 0.24\linewidth]{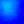}}
    \subfigure{\includegraphics[width = 0.24\linewidth]{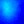}}
    \subfigure{\includegraphics[width = 0.24\linewidth]{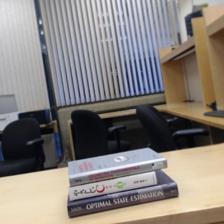}}
    \subfigure{\includegraphics[width = 0.24\linewidth]{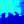}}
    \subfigure{\includegraphics[width = 0.24\linewidth]{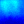}}
    \subfigure{\includegraphics[width = 0.24\linewidth]{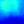}}
    \subfigure[Raw Image]{\includegraphics[width = 0.24\linewidth]{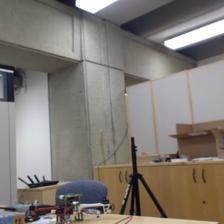}}
    \subfigure[Ground Truth]{\includegraphics[width = 0.24\linewidth]{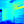}}
    \subfigure[Single Prediction]{\includegraphics[width = 0.24\linewidth]{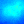}}
    \subfigure[Focal Prediction]{\includegraphics[width = 0.24\linewidth]{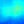}}
    \caption{Testing results on unseen data from our data set comparing prediction from single images and focal stack images. Each row represents an individual testing example. Each column represents the $52^{\mathrm{nd}}$ raw image from focal-stack (far-left), the ground truth depth map (left), the single image prediction (right), and the focal-stack prediction (far-right), respectively.}
\label{fig:05_our_results}
\end{figure}

To compare the methods qualitatively, we illustrate some testing examples on the NYU data set and our own data set in Figs.~\ref{fig:05_nyu_results} and~\ref{fig:05_our_results}, respectively. The NYU testing results (Fig.~\ref{fig:05_nyu_results}) generally fit the ground truth maps, where color here represents the normalized depth. However, it is apparent that there are some major flaws with the prediction. Most notably, the third example of Fig.~\ref{fig:05_nyu_results} does not capture the hallway effect from the ground truth and simply labels the ground and everything else, something that could be improved using focus information. 

The testing results from our data set (Fig.~\ref{fig:05_our_results}) show better results than the NYU data set, especially considering the drastic reduction in training data. Overall, we can see that the estimation using stacked focal images is slightly more accurate and closer to the ground truth than the single image estimation. The focal information also allows us to capture more details and renders more contrast to regions with significant depth difference. For example, in the first example of Fig.~\ref{fig:05_our_results}, the prediction from focus data better identifies the opening of the doorway and its rectangular shape. Similarly, the outline of the computer and desk in the third example is slightly more apparent with the focus data prediction. 

The most important and promising difference in learning on single RGB images compared to the focal stack is the accuracy of the prediction. For example, the testing results in Fig.~\ref{fig:05_our_results} are scaled to their true depth values, i.e., colors in the heat maps represent true distances and not normalized distance values. Therefore, the closer in color the prediction is to the ground truth represents how accurate the predicted depth map is. In almost every testing result (including those presented in Fig.~\ref{fig:05_our_results}), we observed that the depth accuracy from the focal stack is significantly higher than from the single RGB image. For example, the focus result in the sixth example of Fig.~\ref{fig:05_our_results} captures the depth of the foreground and background more accurately than the single image alternative. This is intuitive because the optics of the camera and its focus inherently represent the depth of a scene, which is being captured by the focal stack in our data. 

However, the benefit of the additional focus information is not as apparent as we hoped for several reasons. The primary reason is the lack of data. Even with replication, the number of training examples in our data set is far below the number used in the NYU data set. Secondly, the cheap RGB webcam included a small aperture with limited focal depth, which resulting in larger than desired depth of fields. The result is that the regions of good focus information were too close to the camera ($<0.4m$) and did not overlap well with the Kinect's usable volume. In the future, we will replace this with a higher quality camera with controllable aperture, i.e. a DSLR, allowing us to observe variable focus at father physical distances from the camera and to highlight the true advantages of learning from focus. 

\section{Conclusion and Future Work}
\label{sec:conclusion}


In this project, we successfully implemented a convolutional neural network to learn and predict the pixel-wise depth maps from RGB images. Moreover, we proposed a novel approach which uses a series of out-of-focus images taken with different focal lengths, and we show that this approach outperforms the traditional depth estimation methods using one single RGB image. To validate our idea, we also collected our own data set using customized hardware. To the best of our knowledge, this is the first out-of-focus image data set for depth estimation. Overall, our results show that including several images with variable focus does indeed improve the depth estimation from standard RGB images. 

Although we showed promising results for learning depth from focus, we believe that this project serves primarily as a strong proof of concept. In the future, the problems identified in Section \ref{sec:results} can be addressed by using a high quality camera with better optics specifications designed for small depths-of-field. Additionally, we could gain more insight into this method by collecting more data for training. Increasing the number of images as well as the number of scenes, such as outdoor or cluttered environments, would also improve the results. The neural network could also be improved by tuning hyper parameters or by modifying the current structure to better accommodate the focal stack. 

In the future, we intend on replacing the mobile data collection cart with an autonomous ground robot with the goal of automated data collection. The learned depth estimation could then be utilized by the ground robot to perform navigation, localization, and mapping with real-time collision avoidance. Ultimately, we hope that other researchers utilize image bokeh to build intelligent robots.



\pagebreak
\bibliographystyle{IEEEtran}		
\bibliography{references}

\end{document}